\documentclass[11pt,a4paper]{article}
\usepackage[hyperref]{emnlp2020}
\usepackage{times}
\usepackage{latexsym}

\usepackage{microtype}
\usepackage{graphicx}
\usepackage{booktabs}
\usepackage{adjustbox}
\usepackage{multirow}
\usepackage{makecell}

\usepackage{amssymb}

\AtBeginDocument{}

\aclfinalcopy 

\title{SIGTYP 2020 Shared Task: Prediction of Typological Features}

\newcommand{\aau}{\bullet}
\newcommand{\ku}{\triangleleft}
\newcommand{\ethz}{\star}
\newcommand{\jhu}{\diamond}
\newcommand{\cam}{\ddag}
\newcommand{\ulei}{\odot}
\newcommand{\cmu}{\Box}
\newcommand{\unimelb}{\sharp}

\author{
Johannes Bjerva$^{\aau,\ku}$~
Elizabeth Salesky$^{\jhu}$~
Sabrina J. Mielke$^{\jhu}$~
Aditi Chaudhary$^{\cmu}$ \\
\textbf{
Giuseppe G. A. Celano$^{\ulei}$~
Edoardo M. Ponti$^{\cam}$~
Ekaterina Vylomova$^{\sharp}$
}\\
\textbf{
Ryan Cotterell$^{\ethz}$~
Isabelle Augenstein$^{\ku}$~
}\\
  $^{\aau}$Aalborg University~
  $^{\ku}$University of Copenhagen~
  $^{\jhu}$Johns Hopkins University\\
  $^{\unimelb}$University of Melbourne~
  $^{\cmu}$Carnegie Mellon University~
  $^{\ulei}$Leipzig University\\
  $^{\ethz}$ETH Zürich~
  $^{\cam}$University of Cambridge\\
  \texttt{jbjerva@cs.aau.dk},~\texttt{augenstein@di.ku.dk}
}

\begin{document}
\maketitle

\begin{abstract}
Typological knowledge bases (KBs) such as WALS \cite{wals} contain information about linguistic properties of the world's languages. They have been shown to be useful for downstream applications, including cross-lingual transfer learning and linguistic probing. A major drawback hampering broader adoption of typological KBs is that they are sparsely populated, in the sense that most languages only have annotations for some features, and skewed, in that few features have wide coverage. As typological features often correlate with one another, it is possible to predict them and thus automatically populate typological KBs, which is also the focus of this shared task. 
Overall, the task attracted 8 submissions from 5 teams, out of which the most successful methods make use of such feature correlations.
However, our error analysis reveals that even the strongest submitted systems struggle with predicting feature values for languages where few features are known.
\end{abstract}

\section{Introduction}

Linguistic typology is the study of structural properties of languages \citep{comrie1988linguistic,crofttypology,velupillai2012introduction}. 
Approaches to the categorisation of the languages of the world according to their linguistic properties are represented by, e.g., typological features in databases such as WALS \cite{wals}, URIEL \cite{Littel-et-al:2017}, and AUTOTYP \cite{autotyp}, e.g. in terms of their syntax, morphology, and phonology. 
One example of such a typological feature is the basic word order feature in WALS. For instance, English is best described as a subject-verb-object (SVO) language, whereas Japanese is best described as a subject-object-verb (SOV) language. 

Once a relatively niche topic in the NLP community, studying typological features has recently risen in popularity and importance for a number of reasons. The field has seen considerable advances in cross-lingual transfer learning, whereby stable cross-lingual representations can be learned on massive amounts of data in an unsupervised way, be it for words \cite{journals/corr/AmmarMTLDS16,conf/acl/WadaIM19} or, more recently, sentences \cite{journals/tacl/ArtetxeS19,DBLP:journals/corr/abs-1810-04805,conf/nips/ConneauL19,conf/acl/ConneauKGCWGGOZ20}. This naturally raises the question of what these representations encode, and some have turned to typology for potential answers \cite{journals/corr/abs-2009-12862,zhao2020inducing}. In a similar vein, research has shown that these learned representations can be fine-tuned for supervised tasks, then applied to new languages in a few- or even zero-shot fashion with surprisingly high performance. This has raised the question of what causes such surprisingly high results, and to what degree typological similarities are exploited by such models \cite{bjerva-augenstein-2018-phonology,nooralahzadeh2020meta,zhao2020inducing}.

In addition to using typology for diagnostic purposes, prior work has also found that typology can guide cross-lingual sharing \cite{delhoneux:2018}.
Finally, the relationship between typological knowledge bases (KBs) such as WALS \citep{wals} and language representations has been studied, which has shown that knowledge base population methods can be used to complete typological KBs \cite{malaviya-etal-2017-learning,murawaki:2017,bjerva-augenstein-2018-phonology,bjerva-etal-2019-language}, and that implications can be discovered in typological KBs \cite{daume:2007,bjerva-etal-2019-uncovering}.

\begin{figure*}[t]
    \centering
    \includegraphics[width=\linewidth]{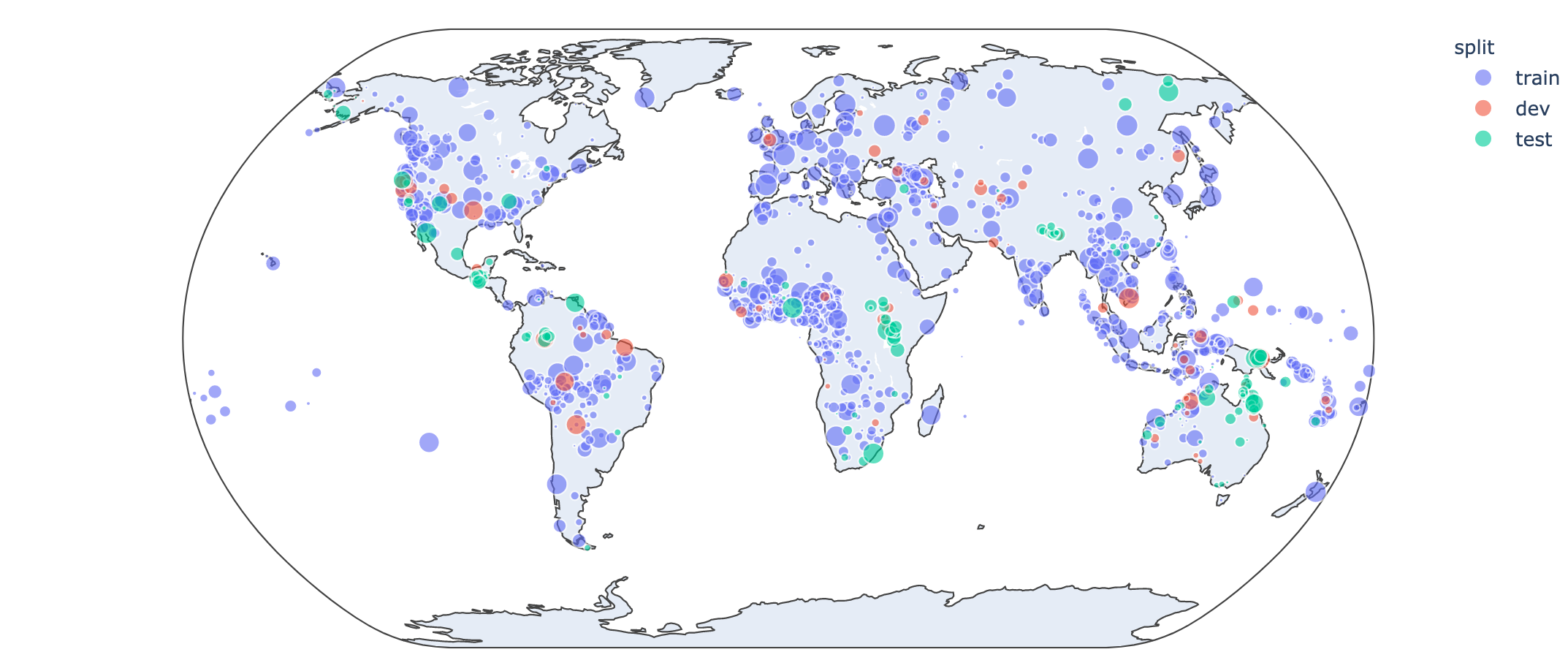}
    \caption{Shared task WALS data superimposed on a map showing one point per language with train, dev, and test splits; relative point sizes representing number of features for that language.}
    \label{fig:data-map}
\end{figure*}

The latter stream of work has provided the inspiration for this shared task on typological feature prediction. As knowledge bases are notoriously incomplete and require manual labour from (in this case, linguistic) domain experts to create, populate and maintain, high-performance methods for automatic knowledge base population are highly desirable. While past approaches have shown the feasibility of typological feature prediction, the considered evaluation setups have some flaws which led to overestimated performance. Some papers control for phylogenetic relationships between languages, e.g. not both training and testing on Slavic languages, but little-to-no work has considered controlling for geographical proximity. This is corrected for in this shared task.

The shared task attracted 8 system submissions from 5 teams for two subtasks (constrained and unconstrained resources). 
In general, the systems which make use of correlations between features, and exploit observed features during inference, perform better, whereas those that do not make use of observed features perform similarly to our baselines.

\section{Task Description}

The SIGTYP 2020 shared task is concerned with predicting typological features from the World Atlas of Language Structures (WALS) \citep{wals}. For the task, participants were invited to build systems to predict features for languages unseen at training time. The shared task consisted of two subtasks: 1) the \textit{constrained} setting, for which only the provided training data may be used; 2) the \textit{unconstrained} setting, for which training data may be extended with any external source of information (e.g. pre-trained embeddings, additional text, etc.) 

\paragraph{Data Format} For each instance, the following information is provided: the language code, name, latitude, longitude, genus, family, country code, and features. At training time, both the feature names and feature values are given, while at test time, submitted systems are required to fill values for the requested features.
An example of a test instance is given in \autoref{tab:data_format}.

\begin{table*}
\setlength{\tabcolsep}{3pt}
\centering
\fontsize{8}{8}\selectfont
\begin{tabular}{ccccccccp{59mm}}
\toprule
& \textbf{Lang code} & \textbf{Name} & \textbf{Lat} & \textbf{Long} & \textbf{Genus} & \textbf{Family} & \textbf{Count Code} & \hfil \textbf{Features} \\ 
\midrule
\multirow{4}{*}{\textbf{Input}} & mhi & Marathi & 19.0 & 76.0 & Indic & Indo-European & IN & \makecell{order\_of\_subject,\_object,\_and\_verb=? \textbar} \makecell{number\_of\_genders=?} \\
\cmidrule(lr){2-9}
&  jpn & Japanese & 37.0 & 140.0 & Japanese & Japanese & JP & \makecell{case\_syncretism=? \textbar} \makecell{order\_of\_adjective\_and\_noun=?} \\
\midrule
\multirow{5}{*}{\textbf{Output}} & mhi & Marathi & 19.0 & 76.0 & Indic & Indo-European & IN & \makecell{order\_of\_subject,\_object,\_and\_verb=SOV \textbar} \makecell{number\_of\_genders=three} \\
\cmidrule(lr){2-9}
&  jpn & Japanese & 37.0 & 140.0 & Japanese & Japanese & JP & \makecell{case\_syncretism=no\_case\_marking \textbar} \makecell{order\_of\_adjective\_and\_noun=demonstrative-Noun} \\
\bottomrule
\end{tabular}
\caption{Data format for two test instances of the SIGTYP 2020 shared task dataset}
\label{tab:data_format}
\end{table*}

\subsection{Dataset}

WALS comprises 2679 languages and a total of 192 feature categories \citep{wals}.
However, the database is quite sparse in that many language-feature combinations lack annotation.
Furthermore, it is a skewed database, in that a handful of languages have annotations for a large number of features, and some features are annotated for almost all languages, whereas some have very little coverage.
In order to alleviate data sparsity in the shared task, only the subset of the languages in WALS with more than 3 features available are considered.
Furthermore, of all the features of the languages so selected only those present in more than 9 languages have been retained.
Most feature categories in WALS can take several feature values.
For instance, the feature \texttt{Tone} can take one of the values: \texttt{No tones}, \texttt{Simple tone system}, or \texttt{Complex tone system}. 
This dataset has been divided into train set (90\%), dev set (5\%), and test set (5\%). 

\section{Evaluation Setup}

While a substantial amount of previous work deals with feature prediction in typological databases such as WALS (e.g.~\citet{malaviya-etal-2017-learning,murawaki:2017,bjerva-augenstein-2018-phonology,bjerva-etal-2019-language}), most such work does not take into account that both phylogenetic and geographic proximity should be controlled for.
Languages which have shared common ancestry will often have similar typological features, hence training and evaluating on the same language family will tend to inflate the expected performance of the model \citep{bjerva-etal-2019-probabilistic}.
In the data for this shared task, we make sure to control for both of these factors.

Our evaluation setup is constructed as follows. 
We evaluate on a set of languages from small languages spread across the world, as defined by the WALS macroareas: Mayan (North America), Tucanoan (South America), Madang (Papuanesia), Mahakiranti (Eurasia), Northern Pama-Nyungan (Australia), and Nilotic (Africa).
In addition, we include a subset of languages spoken around the world, by randomly sampling 10\% of the available data in WALS.
This yields two evaluation set-ups: one in which we evaluate on unobserved languages, controlling for both phylogenetic and geographic relationships, and one in which we perform a random evaluation as is common in previous work.

The languages in the test data vary in the number of removed and present feature values so that the \emph{blanking ratios} are spread uniformly between 5\% and 95\%.
This will allow our analysis to investigate whether some approaches benefit from observing a large number of features and whether some are robust to situations where only a small number of features are observed (\autoref{sec:blankingratios}).

In order to control for phylogenetic and geographic effects, we remove all languages from the same language genus as the aforementioned languages from the training set, as well as all languages which are spoken within 1,000km of any of these languages.\footnote{Distances calculated with WALS language locations.}
This reduces the number of languages in the training set to 1250.
The task had participants run their systems on the partial feature information for our held-out languages and send us the outputs of their systems, i.e., the imputed features.

\subsection{Evaluation Metrics}

We report macro-averaged accuracies, meaning that we first compute the average accuracy for each language, i.e., the fraction of to be imputed features correctly predicted by the participant's system, then average these language accuracies within each language genus, and finally report the average of these genus accuracies to rank participants as well as all these accuracies for each language genus (\autoref{sec:analysis}) to see whether systems behave differently on different language families. We judge statistical significance using a non-parametric two-tailed paired permutation test with 5k samples each.

\subsection{Baselines}
We provide two baselines.
The first is a simple lower-bound baseline based on observing feature frequencies in WALS (Baseline\_frequency in \autoref{fig:rankings}).
For each unobserved feature in the test set, we predict the most frequent feature value from the training set.

The second uses the $k$-nearest neighbours ($k$-NN) algorithm with a simple feature set to predict each unobserved feature, with $k=1$ (Baseline\_knn-imputation in \autoref{fig:rankings}). 
Each language is represented by a language vector ($\vec{l}\in \mathbb{R}^{64}$) trained as a part of a multilingual character-based language model \citep{ostling-tiedemann-2017-continuous}.
During inference, for a language $l$ and unobserved feature $y$, we find the nearest neighbour to $\vec{l}$ for which $y$ has been observed, similar to \citet{,bjerva-augenstein-2018-phonology,bjerva_augenstein:18:iwclul}. 

\subsection{Submissions} 


We received eight submissions from five teams across the \textit{constrained} and \textit{unconstrained} subtasks, as described below. 

\textbf{ÚFAL} (\citet{ufal2020sigtyp}, Charles University) submitted a \textit{constrained} system which ensembled two approaches: first, estimating the correlation of feature values within languages enables missing feature prediction, and second, using a neural network to predict whether feature values match a specific language after training one network with all provided WALS feature values and pre-computed language embeddings. 
By ensembling both using confidence scores, they were able to improve on each individual approach and produce the best accuracy of all constrained and unconstrained submissions. 

\textbf{CrossLingference} (\citet{crosslingference2020sigtyp}, University of Tübingen) submitted an \textit{unconstrained} system using inferred phylogenetic trees. 
These were built with Continuous Time Markov Processes using Swadesh lists from the Automated Similarity Judgment Project (ASJP), with k-nearest neighbour estimations based on geographic information as back-off for test set languages not in both Glottolog and ASJP. 
Ancestral state reconstruction allows the inference of features for ancestral states from the provided surface features (WALS), and similarly, for this year's shared task, unknown feature values for non-ancestral languages can be inferred individually by rerooting the tree to a related language. 

\textbf{NUIG} (\citet{nuig2020sigtyp}, NUI Galway) submitted a \textit{constrained} system with independent classifiers to predict each WALS feature. The outputs of independent classifiers are then fed into a shared encoder with feed-forward and self-attention layers in order to make use of feature correlations. 
Their model does not use other known features for WALS feature prediction at inference time, relying only on the 5-dimensional inputs of longitude, latitude, genus, family, and country-code. 

\textbf{NEMO} (\citet{nemo2020sigtyp}, Google London and Tokyo) submitted \textit{constrained} systems which first computed probabilities of represented feature values across each language's genetic (genus and family), and areal (features from languages within a 2,500 kilometre radius, computed from provided latitude and longitude with the Haversine formula), and \textit{implicational universals} or rather, priors for certain features given commonly associated feature-value pairs in the data. They compared several classifiers' performance using these sparse features, ultimately submitting systems using ridge regression. 
The two submitted systems differ in whether these features were computed for the test set or only train and dev. 

\textbf{Panlingua} \cite{panlingua2020sigtyp}, a team effort across KMI, Panlingua, and IIT KGP, submitted \textit{constrained} systems from three approaches: two rule-based systems (one statistical, and one frequency-based baseline) and one hybrid system. 
Their baseline is similar to the organizers' frequency-base baseline, except that it produces the most frequent value for a feature within a genus if available, backing off to language family, and then the overall most-frequent value. 
The hybrid system uses 180 different SVM classifiers for the 180 features which were present in the training set. 
The statistical system provides a two-step back off procedure if neither a feature has been seen for either a languages' genus or family in training: first, finding the most frequent values in nearby languages using Haversine distance, and if these are too distant, turning to nearby language families. This system performed best on the held-out data. 

\section{Results}
\label{sec:results}

\begin{figure}[!bt]
    \centering
    \includegraphics[width=\columnwidth]{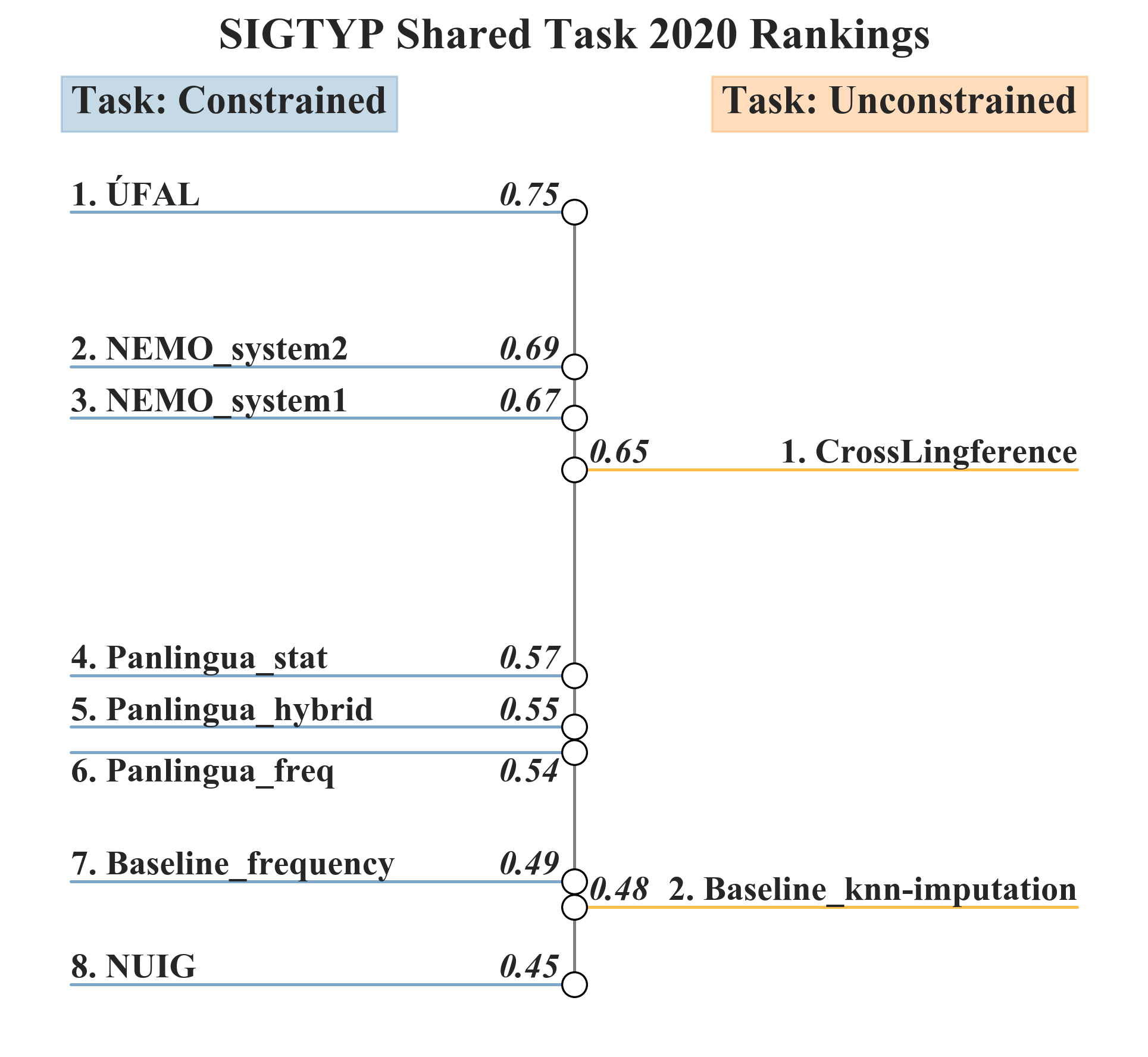}
    \caption{Macro-averaged rankings of all submissions}
    \label{fig:rankings}
\end{figure}

\begin{figure}[t]
    \centering
    \includegraphics[width=\columnwidth]{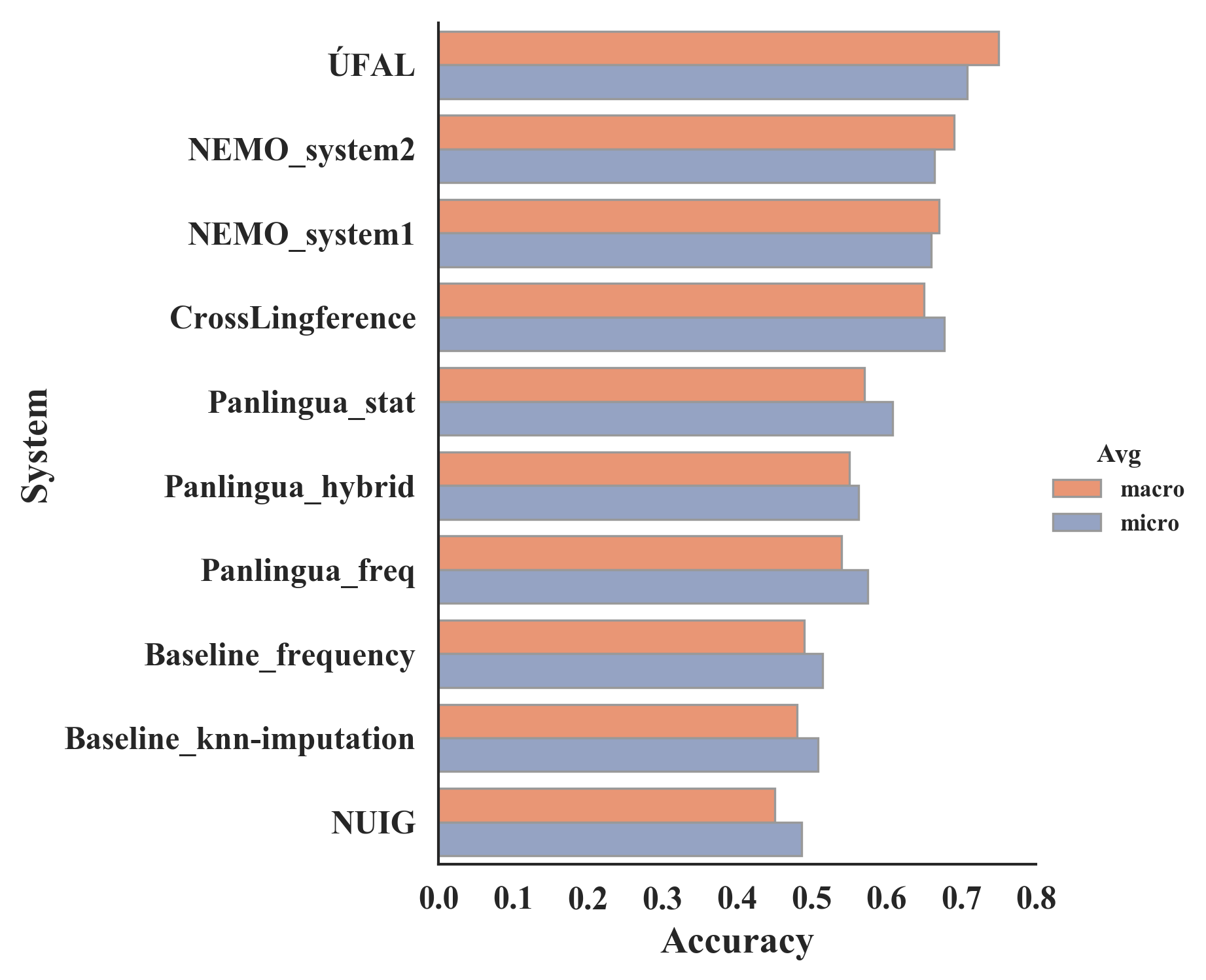}
    \caption{Comparison of macro-averaged and micro-averaged accuracies across submissions}
    \label{fig:macro-micro}
\end{figure}

\autoref{fig:rankings} shows the overall results and rankings for all shared task submissions. 
The rankings use macro-averaged accuracies as this equally weights the controlled genera (the exception is the comparison to micro-averaged accuracies in \autoref{fig:macro-micro}).
This year's shared task was separated into two subtasks: \textit{constrained} systems which used only the WALS features and data provided, and \textit{unconstrained} systems, open to use of any data or pre-trained models. 
Accordingly, we have two winning systems: \textbf{ÚFAL} for \textit{constrained}, and \textbf{CrossLingference} for \textit{unconstrained}, with \textbf{ÚFAL} producing the best results overall across both subtasks.

Results for each unobserved genus, shown in comparison to results across genera observed in training, may be found in \autoref{table:macro-genera}. 

WALS feature value formatting is not standardized and, unfortunately, the test data was released containing additional tabs within the feature values for some features, which adversely affected teams who may have used tab-separation for data preprocessing. 
Many teams accounted for this and submitted feature values for all 2417 features across the 149 languages in the test data, but for two teams this led to missing features in their submissions:  \textbf{CrossLingference} was missing 7 features across 7 languages, affecting their results by $1\%$; \textbf{Panlingua} was missing 61 features across 15 languages in their rule-based submission and 57 across 11 for their additional two submissions, affecting their results by $2\%$. 
When evaluating without the affected features, rankings were not changed, nor were there significant differences between submitted systems. 


\subsection{Subtask 1: Constrained Setting}

The nine systems in the constrained setting used a diverse set of model features and architectures. 
When computing pairwise significances with a paired permutation test, we find that these systems cluster into three groups, within each the systems are not significantly different from each other: \{\textbf{1}\}, \{\textbf{2-3}\}, and \{\textbf{4-8}\}.
Teams submitting multiple systems were able to improve their accuracy within their own submissions, but we did not find that their individual submissions were statistically significantly different.
Similar differences in overall accuracy do not necessarily indicate statistically significant margins: for example, the 1st and 2nd systems have the same margin (0.05) as the 4th and 7th, but the latter are not significantly different while the former are. 

Finer-grained analysis of results across controlled genera, and comparing results across different levels of representation in the training data, can be found in \autoref{sec:analysis}. 

\subsection{Subtask 2: Unconstrained Setting}

\textbf{CrossLingference} submitted the only unconstrained system, which used additional data in the form of Swadesh lists to infer phylogenetic trees. This system outperforms the unconstrained knn-imputation baseline on all evaluated conditions. 
When we contextualize this submission by comparing it to those in the constrained setting, we find that it joins the second cluster with the two submissions from \textbf{NEMO}; interestingly, when features are micro-averaged rather than macro-averaged, these teams reorder, with \textbf{CrossLingference} outperforming the two \textbf{NEMO} systems, seen in \autoref{fig:macro-micro}. 
This is somewhat counter-intuitive, given the way each system uses phylogenetic information. 
While \textbf{CrossLingference} explicitly models phylogenetic information through its model structure, \textbf{NEMO} takes a frequentist approach where the counts and probabilities of each feature within a language's genus, family, and geographic area are pre-computed and passed as sparse features to feature classifiers. 
One might expect the latter to perform better on a micro-average where overall data frequencies would be more heavily weighted than each genus, but this was not the case here. 
We explore this further in \autoref{sec:analysis}.

\section{Analysis}
\label{sec:analysis}

\begin{table*}[tb]
\begin{adjustbox}{width=\linewidth}
\begin{tabular}{lccccccc}
\toprule
\multirow{2}{*}{\textbf{Submission}} & \textbf{Tucanoan} & \textbf{Madang} & \textbf{Mahakiranti} & \textbf{Nilotic} & \textbf{Mayan} & \textbf{N. Pama-Nyungan} & \textbf{Other genera} \\
  & (8)               & (9)             & (13)                 & (15)             & (17)           & (24)                     & (63)                  \\
\cmidrule(lr){1-1} \cmidrule(lr){2-2} \cmidrule(lr){3-3} \cmidrule(lr){4-4} \cmidrule(lr){5-5} \cmidrule(lr){6-6} \cmidrule(lr){7-7} \cmidrule(lr){8-8}
ÚFAL                         & \textbf{0.73}     & \textbf{0.78}   & \textbf{0.74}        & 0.71    & \textbf{0.80}  & \textbf{0.76}                  & \textbf{0.76}         \\
NEMO\_system2                & 0.71     & 0.72   & 0.72        & \textbf{0.76}    & 0.76  & 0.67                  & 0.69         \\
NEMO\_system1                & 0.70     & 0.72   & 0.68        & 0.75    & 0.71  & 0.68                  & 0.67         \\
Panlingua\_stat              & 0.70     & 0.64   & 0.55        & 0.55    & 0.33  & 0.62                  & 0.58         \\
Panlingua\_hybrid            & 0.65     & 0.64   & 0.57        & 0.51    & 0.34  & 0.61                  & 0.53         \\
Panlingua\_freq              & 0.59     & 0.64   & 0.53        & 0.55    & 0.31  & 0.59                  & 0.55         \\
\textit{Baseline\_frequency} & 0.51     & 0.53   & 0.37        & 0.49    & 0.41  & 0.58                  & 0.53         \\
NUIG                         & 0.51     & 0.56   & 0.35        & 0.45    & 0.32  & 0.45                  & 0.48         \\
\midrule
CrossLingference             & 0.71      & 0.73   & 0.67        & 0.68    & 0.57  & 0.60                  & 0.65         \\
\textit{Baseline\_knn-imputation} & 0.48 & 0.57   & 0.46        & 0.48    & 0.32  & 0.52                  & 0.51        \\
\bottomrule
\end{tabular}
\end{adjustbox}
\caption{Macro-averaged results across each unobserved genus, as compared to genera with languages observed in training with randomly sampled splits, shown with number of languages in each genus. }
\label{table:macro-genera}
\end{table*}

Our test data was constructed to enable comparison across controlled phylogenetic and geographic relationships, and randomly sampled features from languages covered in training as is common in previous work.

\subsection{Overall Results}

\autoref{table:macro-genera} compares submission accuracy on features from diverse WALS macroareas unobserved in training data, and other observed genera. 
We see that overall rankings hold when evaluated on observed languages.
However, this is not the case for several of our unobserved genera. 
With respect to the shared task baselines, we find that the frequency baseline, which naively picks the most well-represented values for each feature, is most representative for the larger `other genera' category, which represents the majority of the training data but does not account for the diversity of typological features and values across many languages. 
Nonetheless, for most of the unobserved genera, the frequency baseline performed better than the knn-imputation baseline, which was significantly better for Mahakiranti only, primarily due to correct prediction of ``OV'' ordering across multiple features.

Interestingly, while the first 3 systems perform better on macro-averaged accuracy than micro-averaged (\autoref{fig:macro-micro}), this is not true for all other systems, suggesting that they rely more on getting frequent and ``easy'' features right, relying on frequency in training data. Note that the six unobserved genera come from separate macroareas across six different continents, and have a more even distribution of feature values than the `other genera.'

\subsection{Differences across Genera}

Looking at specific genera, we see that Mayan caused the greatest split between submitted systems, with the first two clusters performing very well, and the frequency baseline performed better than the majority of systems. 
On the other end of the spectrum, certain genera (Tucanoan and Madang) with well-represented features were relative equalizers, with the least variance in results across the submitted systems.

Within those teams which submitted multiple systems, there were only certain cases in which these performed significantly differently from each other. 
\textbf{Panlingua} submitted three different systems; two rule-based (one statistical and one frequency-based), and one hybrid model. 
For most genera, these performed very similarly, with consistently better results from the statistical rule-based system than the others, though there were no statistically significant differences shown by paired permutation tests. However, this was not the case for Tucanoan, where the statistical (and to a less degree, hybrid) model significantly outperformed the other. 
These systems had equal performance on four of the Tucanoan languages \{Cubeo,Secoya,Siona,Koreguaje\}, but quite divergent on the remaining four languages \{Desano,Retuarã,Tucano,Tuyuca\}. This second set required predicting values for several features concerning the order of Subject, Object, Verb, which the statistical model was able to correctly predict through better back-off choices, but swayed by the more frequent SVO languages in training, their frequency baseline and SVM-based classifiers were not. 


\subsection{Differences among Features}

\begin{table*}[t]
\begin{adjustbox}{width=\linewidth}
\begin{tabular}{llccc}
\toprule
                  & Feature    & \# Langs                                                              & \multicolumn{1}{l}{Avg. Accuracy} & \multicolumn{1}{l}{Std. Deviation} \\ \midrule
\multirow{5}{*}{Highest} & front\_rounded\_vowels & 4                                                  & 0.65                                 & 0.08                               \\
                         & inclusive/exclusive\_forms\_in\_pama-nyungan & 2                           & 0.65                                 & 0.09                               \\
                         & distributive\_numerals & 3                                                  & 0.64                                 & 0.08                               \\
                         & optional\_double\_negation\_in\_svo\_languages & 1                                  & 0.64                                 & 0.08                               \\
                         & voicing\_in\_plosives\_and\_fricatives & 4                                                           & 0.63                                 & 0.08                               \\
\midrule
\multirow{5}{*}{Lowest}  & verb-initial\_with\_clause-final\_negative & 1                              & 0.44                                 & 0.21                               \\ 
                         & multiple\_negative\_constructions\_in\_svo\_languages & 3                   & 0.41                                 & 0.27                               \\
                         & suppletion\_in\_imperatives\_and\_hortatives &  2                           & 0.40                                 & 0.13                               \\
                         & languages\_with\_two\_dominant\_orders\_of\_subject,\_object,\_and\_verb & 2 & 0.39                                 & 0.20                               \\
                         & the\_position\_of\_negative\_morphemes\_in\_verb-initial\_languages & 9     & 0.38                                 & 0.28                               \\
                         \bottomrule
\end{tabular}
\end{adjustbox}
\caption{Features with the highest and lowest overall accuracies across \textit{all} submissions, with number of languages containing the feature in the test data (183 total languages)}
\label{tab:high-low-accuracies}
\end{table*}

\autoref{tab:high-low-accuracies}
shows the features with the highest and lowest accuracies across all submissions. 
We find that the features with highest accuracies also have the most consistent performance across all systems, and typically have the most frequent values for those features. 
The most difficult features, on the other hand, have the least frequently occurring values in the training data, and have higher variance -- interestingly, the top four systems were nonetheless able to achieve greater than 65\% accuracy on these features, while the remaining systems' accuracies were $\sim20\%$.

\subsection{Impact of Blanking Ratio}\label{sec:blankingratios}

Since the languages in our test sets all remove and retain different numbers of features, we can see whether the blanking ratio, that is, the ratio of features that participants have to impute to the features listed for that language, correlates with the performance of the system in question on that language. Calculating Pearson's correlation coefficient for every system individually, we realize that they range from -0.23 (for a NEMO system) to 0.31 (for a Panlingua system), almost all of which statistically significantly different from 0.

\begin{figure}[!bt]
    \centering
    \includegraphics[width=\columnwidth]{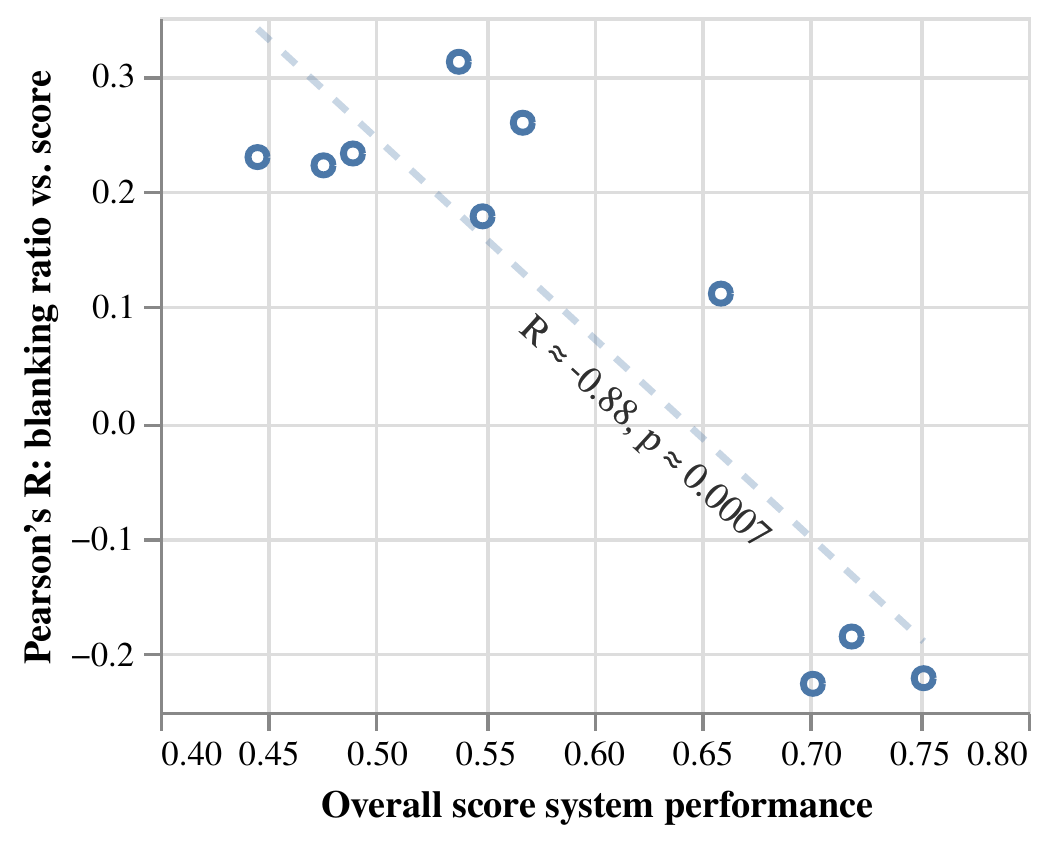}
    \caption{Correlation coefficients (between blanking ratio and language performance) for each system as a function of that system's performance. The correlation of $\mathrm{R}=-0.88$ is significant at $p < .001$.}
    \label{fig:blankingratios}
\end{figure}

Why do these correlations differ so much from system to system? To answer that question, we plot these correlation coefficients as a function of the system's overall performance in \autoref{fig:blankingratios}.
It turns out that a system's overall performance and how sensitive it is to the blanking ratio are highly correlated: the stronger systems are much more negatively affected by the removal of more features (their correlations are negative), weaker systems are not only not harmed, but seem to find the languages where only a few features are blanked harder still (having positive correlations).

\section{Related Work}

Previous work can be divided into research on predicting typological features automatically, cross-lingual transfer learning which utilises typology to inform sharing, probing of representations for what typological knowledge they encode, and finally, work on how best to represent a language in terms of its typological features.

\subsection{Predicting Typological Features}
Typological knowledge bases are both sparse and skewed in terms of language--feature annotations. They are sparse in the sense that most languages only have annotations for a handful of features and skewed in the sense that a few features have much wider coverage than others.
Luckily, such features often correlate with one another, which allows for prediction of those features from others. For instance, languages where the verb precedes the object tend to have prepositions, e.g. Norwegian, whereas languages where the object precedes the verb word tend to have postpositions, e.g. Japanese. 

A survey of approaches to prediction of features is provided in \citet[\S~4.3]{ponti2019modeling}. 
Some common approaches include prediction based on language representations learned as a by-product of model training \citep{ostling-tiedemann-2017-continuous,malaviya-etal-2017-learning,bjerva-augenstein-2018-phonology,bjerva-etal-2019-language} and matrix factorisation \citep{murawaki:2017,bjerva-etal-2019-probabilistic}.

\subsection{Typologically Informed Sharing}
Cross-lingual sharing informed by typology has been investigated for, among others, parsing \cite{naseem-etal-2012-selective,tackstrom-etal-2013-target,zhang-barzilay-2015-hierarchical,delhoneux:2018}, language modeling \citep{tsvetkov-etal-2016-polyglot,ponti-etal-2019-towards}, machine translation \citep{daiber-etal-2016-universal,ponti-etal-2018-isomorphic}, and morphological inflection \citep{chaudhary-etal-2019-cmu}. 
Many of these approaches use language embeddings with sparse features encoding WALS feature values. 
\citet{oncevay2020bridging} find that combining information from typological databases with embeddings learned during training of an NMT model can be beneficial for multilingual NMT.

\subsection{Typological Probing}
Several recent papers study typological feature prediction as a probing task for evaluating cross-lingual sentence encoders \cite{journals/corr/abs-2009-12862,bjerva-augenstein-2018-phonology,nooralahzadeh2020meta,zhao2020inducing}.
Typically, hidden representations are probed for whether or not they might encode a typological feature by, e.g., using them in a separate classifier \citep{malaviya-etal-2017-learning,bjerva-augenstein-2018-phonology,nooralahzadeh2020meta}.
\citet{ostling-tiedemann-2017-continuous} learn language representations during multilingual language modelling and find that the resulting representations can reproduce relatively credible phylogenetic trees.

\citet{bjerva-augenstein-2018-phonology} learn language representations under NLP tasks such as POS tagging and grapheme-to-phoneme conversion, and find that typological features related to the task at hand are sometimes encoded.
\citet{nooralahzadeh2020meta} use a typological probing task in experiments for zero- and few-shot NLI and QA, finding that languages which share typological properties benefit from sharing.  
\citet{zhao2020inducing} attempt to induce language-agnostic representations, e.g.~by reducing the typological gaps between languages, and find that this is beneficial for NLI and MT.
\citet{gerz-etal-2018-relation} show that there is a correlation between typological features related to morphology and model performance in language modelling, and \citet{cotterell-etal-2018-languages-v2} further show that inflectional morphology affects performance in both $n$-gram and LSTM-based language models.


\section{Conclusions}

This paper documents the first SIGTYP shared task on prediction of typological features in WALS.
The 8 system submissions from 5 teams showed that a variety of different methods can be applied to the task.
Interestingly, the best system only achieved a macro-averaged accuracy of 75\%, indicating that the task is far from solved.
This further shows that the evaluation set-up in which we controlled for both phylogenetic relationships and geographic proximity is a challenging one.
We expect that further exploration of unconstrained systems to have the most potential for predicting features in such cases, where little or nothing is known about a language.


\section*{Acknowledgments}
This research has received funding from the Swedish Research Council under grant agreement No 2019-04129, as well as
the German Research Foundation (DFG project number 408121292).

\bibliography{anthology,emnlp2020}

\begin{thebibliography}{42}
\expandafter\ifx\csname natexlab\endcsname\relax\def\natexlab#1{#1}\fi

\bibitem[{Ammar et~al.(2016)Ammar, Mulcaire, Tsvetkov, Lample, Dyer, and
  Smith}]{journals/corr/AmmarMTLDS16}
Waleed Ammar, George Mulcaire, Yulia Tsvetkov, Guillaume Lample, Chris Dyer,
  and Noah~A. Smith. 2016.
\newblock \href
  {http://dblp.uni-trier.de/db/journals/corr/corr1602.html#AmmarMTLDS16}
  {{Massively Multilingual Word Embeddings}}.
\newblock \emph{CoRR}, abs/1602.01925.

\bibitem[{Artetxe and Schwenk(2019)}]{journals/tacl/ArtetxeS19}
Mikel Artetxe and Holger Schwenk. 2019.
\newblock \href
  {http://dblp.uni-trier.de/db/journals/tacl/tacl7.html#ArtetxeS19} {{Massively
  Multilingual Sentence Embeddings for Zero-Shot Cross-Lingual Transfer and
  Beyond}}.
\newblock \emph{Trans. Assoc. Comput. Linguistics}, 7:597--610.

\bibitem[{Bjerva and
  Augenstein(2018{\natexlab{a}})}]{bjerva-augenstein-2018-phonology}
Johannes Bjerva and Isabelle Augenstein. 2018{\natexlab{a}}.
\newblock \href {https://doi.org/10.18653/v1/N18-1083} {From phonology to
  syntax: Unsupervised linguistic typology at different levels with language
  embeddings}.
\newblock In \emph{Proceedings of the 2018 Conference of the North {A}merican
  Chapter of the Association for Computational Linguistics: Human Language
  Technologies, Volume 1 (Long Papers)}, pages 907--916, New Orleans,
  Louisiana. Association for Computational Linguistics.

\bibitem[{Bjerva and
  Augenstein(2018{\natexlab{b}})}]{bjerva_augenstein:18:iwclul}
Johannes Bjerva and Isabelle Augenstein. 2018{\natexlab{b}}.
\newblock \href {https://doi.org/10.18653/v1/W18-0207} {{Tracking Typological
  Traits of Uralic Languages in Distributed Language Representations}}.
\newblock In \emph{Proceedings of the Fourth International Workshop on
  Computational Linguistics of Uralic Languages}, pages 76--86, Helsinki,
  Finland. Association for Computational Linguistics.

\bibitem[{Bjerva et~al.(2019{\natexlab{a}})Bjerva, Kementchedjhieva, Cotterell,
  and Augenstein}]{bjerva-etal-2019-probabilistic}
Johannes Bjerva, Yova Kementchedjhieva, Ryan Cotterell, and Isabelle
  Augenstein. 2019{\natexlab{a}}.
\newblock \href {https://doi.org/10.18653/v1/N19-1156} {A probabilistic
  generative model of linguistic typology}.
\newblock In \emph{Proceedings of the 2019 Conference of the North {A}merican
  Chapter of the Association for Computational Linguistics: Human Language
  Technologies, Volume 1 (Long and Short Papers)}, pages 1529--1540,
  Minneapolis, Minnesota. Association for Computational Linguistics.

\bibitem[{Bjerva et~al.(2019{\natexlab{b}})Bjerva, Kementchedjhieva, Cotterell,
  and Augenstein}]{bjerva-etal-2019-uncovering}
Johannes Bjerva, Yova Kementchedjhieva, Ryan Cotterell, and Isabelle
  Augenstein. 2019{\natexlab{b}}.
\newblock \href {https://doi.org/10.18653/v1/P19-1382} {Uncovering
  probabilistic implications in typological knowledge bases}.
\newblock In \emph{Proceedings of the 57th Annual Meeting of the Association
  for Computational Linguistics}, pages 3924--3930, Florence, Italy.
  Association for Computational Linguistics.

\bibitem[{Bjerva et~al.(2019{\natexlab{c}})Bjerva, {\"O}stling, Veiga,
  Tiedemann, and Augenstein}]{bjerva-etal-2019-language}
Johannes Bjerva, Robert {\"O}stling, Maria~Han Veiga, J{\"o}rg Tiedemann, and
  Isabelle Augenstein. 2019{\natexlab{c}}.
\newblock \href {https://doi.org/10.1162/coli_a_00351} {What do language
  representations really represent?}
\newblock \emph{Computational Linguistics}, 45(2):381--389.

\bibitem[{Chaudhary et~al.(2019)Chaudhary, Salesky, Bhat, Mortensen, Carbonell,
  and Tsvetkov}]{chaudhary-etal-2019-cmu}
Aditi Chaudhary, Elizabeth Salesky, Gayatri Bhat, David~R. Mortensen, Jaime
  Carbonell, and Yulia Tsvetkov. 2019.
\newblock \href {https://doi.org/10.18653/v1/W19-4208} {{CMU}-01 at the
  {SIGMORPHON} 2019 shared task on crosslinguality and context in morphology}.
\newblock In \emph{Proceedings of the 16th Workshop on Computational Research
  in Phonetics, Phonology, and Morphology}, pages 57--70, Florence, Italy.
  Association for Computational Linguistics.

\bibitem[{Choenni and Shutova(2020)}]{journals/corr/abs-2009-12862}
Rochelle Choenni and Ekaterina Shutova. 2020.
\newblock \href
  {http://dblp.uni-trier.de/db/journals/corr/corr2009.html#abs-2009-12862}
  {What does it mean to be language-agnostic? probing multilingual sentence
  encoders for typological properties.}
\newblock \emph{CoRR}, abs/2009.12862.

\bibitem[{Choudhary(2020)}]{nuig2020sigtyp}
Chinmay Choudhary. 2020.
\newblock {NUIG: Multitasking Self-attention based approach to SigTyp 2020
  Shared Task}.
\newblock In \emph{Proceedings of the Second Workshop on Computational Research
  in Linguistic Typology}. Association for Computational Linguistics.

\bibitem[{Comrie(1988)}]{comrie1988linguistic}
Bernard Comrie. 1988.
\newblock Linguistic typology.
\newblock \emph{Annual Review of Anthropology}, 17:145--159.

\bibitem[{Conneau et~al.(2020)Conneau, Khandelwal, Goyal, Chaudhary, Wenzek,
  Guzmán, Grave, Ott, Zettlemoyer, and Stoyanov}]{conf/acl/ConneauKGCWGGOZ20}
Alexis Conneau, Kartikay Khandelwal, Naman Goyal, Vishrav Chaudhary, Guillaume
  Wenzek, Francisco Guzmán, Edouard Grave, Myle Ott, Luke Zettlemoyer, and
  Veselin Stoyanov. 2020.
\newblock \href
  {http://dblp.uni-trier.de/db/conf/acl/acl2020.html#ConneauKGCWGGOZ20}
  {{Unsupervised Cross-lingual Representation Learning at Scale}}.
\newblock In \emph{ACL}, pages 8440--8451. Association for Computational
  Linguistics.

\bibitem[{Conneau and Lample(2019)}]{conf/nips/ConneauL19}
Alexis Conneau and Guillaume Lample. 2019.
\newblock \href
  {http://dblp.uni-trier.de/db/conf/nips/nips2019.html#ConneauL19}
  {{Cross-lingual Language Model Pretraining}}.
\newblock In \emph{NeurIPS}, pages 7057--7067.

\bibitem[{Cotterell et~al.(2018)Cotterell, Mielke, Eisner, and
  Roark}]{cotterell-etal-2018-languages-v2}
Ryan Cotterell, Sabrina~J. Mielke, Jason Eisner, and Brian Roark. 2018.
\newblock \href {https://doi.org/10.18653/v1/N18-2085} {Are all languages
  equally hard to language-model?}
\newblock In \emph{Proceedings of the 2018 Conference of the North {A}merican
  Chapter of the Association for Computational Linguistics: Human Language
  Technologies, Volume 2 (Short Papers)}, pages 536--541, New Orleans,
  Louisiana. Association for Computational Linguistics.

\bibitem[{Croft(2002)}]{crofttypology}
William Croft. 2002.
\newblock \emph{{Typology and Universals}}.
\newblock Cambridge University Press.

\bibitem[{Daiber et~al.(2016)Daiber, Stanojevi{\'c}, and
  Sima{'}an}]{daiber-etal-2016-universal}
Joachim Daiber, Milo{\v{s}} Stanojevi{\'c}, and Khalil Sima{'}an. 2016.
\newblock \href {https://www.aclweb.org/anthology/C16-1298} {Universal
  reordering via linguistic typology}.
\newblock In \emph{Proceedings of {COLING} 2016, the 26th International
  Conference on Computational Linguistics: Technical Papers}, pages 3167--3176,
  Osaka, Japan. The COLING 2016 Organizing Committee.

\bibitem[{{Daum\'e III} and Campbell(2007)}]{daume:2007}
Hal {Daum\'e III} and Lyle Campbell. 2007.
\newblock \href {https://www.aclweb.org/anthology/P07-1009} {{A Bayesian Model
  for Discovering Typological Implications}}.
\newblock In \emph{Proceedings of the 45th Annual Meeting of the Association of
  Computational Linguistics}, pages 65--72, Prague, Czech Republic. Association
  for Computational Linguistics.

\bibitem[{Devlin et~al.(2019)Devlin, Chang, Lee, and
  Toutanova}]{DBLP:journals/corr/abs-1810-04805}
Jacob Devlin, Ming-Wei Chang, Kenton Lee, and Kristina Toutanova. 2019.
\newblock \href {https://doi.org/10.18653/v1/N19-1423} {{BERT: Pre-training of
  Deep Bidirectional Transformers for Language Understanding}}.
\newblock In \emph{Proceedings of the 2019 Conference of the North {A}merican
  Chapter of the Association for Computational Linguistics: Human Language
  Technologies, Volume 1 (Long and Short Papers)}, pages 4171--4186,
  Minneapolis, Minnesota. Association for Computational Linguistics.

\bibitem[{Dryer and Haspelmath(2013)}]{wals}
Matthew~S. Dryer and Martin Haspelmath, editors. 2013.
\newblock \href {https://wals.info/} {\emph{WALS Online}}.
\newblock Max Planck Institute for Evolutionary Anthropology, Leipzig.

\bibitem[{Gerz et~al.(2018)Gerz, Vuli{\'c}, Ponti, Reichart, and
  Korhonen}]{gerz-etal-2018-relation}
Daniela Gerz, Ivan Vuli{\'c}, Edoardo~Maria Ponti, Roi Reichart, and Anna
  Korhonen. 2018.
\newblock \href {https://doi.org/10.18653/v1/D18-1029} {On the relation between
  linguistic typology and (limitations of) multilingual language modeling}.
\newblock In \emph{Proceedings of the 2018 Conference on Empirical Methods in
  Natural Language Processing}, pages 316--327, Brussels, Belgium. Association
  for Computational Linguistics.

\bibitem[{Gutkin and Sproat(2020)}]{nemo2020sigtyp}
Alexander Gutkin and Richard Sproat. 2020.
\newblock {NEMO: Frequentist Inference Approach to Constrained Linguistic
  Typology Feature Prediction in SIGTYP 2020 Shared Task}.
\newblock In \emph{Proceedings of the Second Workshop on Computational Research
  in Linguistic Typology}. Association for Computational Linguistics.

\bibitem[{J{\"a}ger(2020)}]{crosslingference2020sigtyp}
Gerhard J{\"a}ger. 2020.
\newblock Imputing typological values via phylogenetic inference.
\newblock In \emph{Proceedings of the Second Workshop on Computational Research
  in Linguistic Typology}. Association for Computational Linguistics.

\bibitem[{Kumar et~al.(2020)Kumar, Alok, Bansal, Lahiri, and
  Ojha}]{panlingua2020sigtyp}
Ritesh Kumar, Deepak Alok, Akanksha Bansal, Bornini Lahiri, and Atul~Kr. Ojha.
  2020.
\newblock {KMI-Panlingua-IITKGP at SIGTYP2020: Exploring rules and hybrid
  systems for automatic prediction of typological features}.
\newblock In \emph{Proceedings of the Second Workshop on Computational Research
  in Linguistic Typology}. Association for Computational Linguistics.

\bibitem[{de~Lhoneux et~al.(2018)de~Lhoneux, Bjerva, Augenstein, and
  S{\o}gaard}]{delhoneux:2018}
Miryam de~Lhoneux, Johannes Bjerva, Isabelle Augenstein, and Anders S{\o}gaard.
  2018.
\newblock \href {https://doi.org/10.18653/v1/D18-1543} {Parameter sharing
  between dependency parsers for related languages}.
\newblock In \emph{Proceedings of the 2018 Conference on Empirical Methods in
  Natural Language Processing}, pages 4992--4997, Brussels, Belgium.
  Association for Computational Linguistics.

\bibitem[{Littell et~al.(2017)Littell, Mortensen, Lin, Kairis, Turner, and
  Levin}]{Littel-et-al:2017}
Patrick Littell, David~R. Mortensen, Ke~Lin, Katherine Kairis, Carlisle Turner,
  and Lori Levin. 2017.
\newblock \href {http://aclweb.org/anthology/E17-2002} {Uriel and lang2vec:
  Representing languages as typological, geographical, and phylogenetic
  vectors}.
\newblock In \emph{Proceedings of the 15th Conference of the European Chapter
  of the Association for Computational Linguistics: Volume 2, Short Papers},
  pages 8--14. Association for Computational Linguistics.

\bibitem[{Malaviya et~al.(2017)Malaviya, Neubig, and
  Littell}]{malaviya-etal-2017-learning}
Chaitanya Malaviya, Graham Neubig, and Patrick Littell. 2017.
\newblock \href {https://doi.org/10.18653/v1/D17-1268} {Learning language
  representations for typology prediction}.
\newblock In \emph{Proceedings of the 2017 Conference on Empirical Methods in
  Natural Language Processing}, pages 2529--2535, Copenhagen, Denmark.
  Association for Computational Linguistics.

\bibitem[{Murawaki(2017)}]{murawaki:2017}
Yugo Murawaki. 2017.
\newblock \href {https://www.aclweb.org/anthology/I17-1046} {Diachrony-aware
  induction of binary latent representations from typological features}.
\newblock In \emph{Proceedings of the Eighth International Joint Conference on
  Natural Language Processing (Volume 1: Long Papers)}, pages 451--461, Taipei,
  Taiwan. Asian Federation of Natural Language Processing.

\bibitem[{Naseem et~al.(2012)Naseem, Barzilay, and
  Globerson}]{naseem-etal-2012-selective}
Tahira Naseem, Regina Barzilay, and Amir Globerson. 2012.
\newblock \href {https://www.aclweb.org/anthology/P12-1066} {Selective sharing
  for multilingual dependency parsing}.
\newblock In \emph{Proceedings of the 50th Annual Meeting of the Association
  for Computational Linguistics (Volume 1: Long Papers)}, pages 629--637, Jeju
  Island, Korea. Association for Computational Linguistics.

\bibitem[{Nichols et~al.(2013)Nichols, Witzlack-Makarevich, and
  Bickel}]{autotyp}
Johanna Nichols, Alena Witzlack-Makarevich, and Balthasar Bickel. 2013.
\newblock The autotyp genealogy and geography database: 2013 release.
\newblock \emph{Zurich: University of Zurich}.

\bibitem[{Nooralahzadeh et~al.(2020)Nooralahzadeh, Bekoulis, Bjerva, and
  Augenstein}]{nooralahzadeh2020meta}
Farhad Nooralahzadeh, Giannis Bekoulis, Johannes Bjerva, and Isabelle
  Augenstein. 2020.
\newblock {Zero-Shot Cross-Lingual Transfer with Meta Learning}.
\newblock In \emph{Proceedings of EMNLP}. Association for Computational
  Linguistics.

\bibitem[{Oncevay et~al.(2020)Oncevay, Haddow, and Birch}]{oncevay2020bridging}
Arturo Oncevay, Barry Haddow, and Alexandra Birch. 2020.
\newblock Bridging linguistic typology and multilingual machine translation
  with multi-view language representations.
\newblock In \emph{Proceedings of EMNLP}. Association for Computational
  Linguistics.
\newblock ArXiv preprint arXiv:2004.14923.

\bibitem[{{\"O}stling and Tiedemann(2017)}]{ostling-tiedemann-2017-continuous}
Robert {\"O}stling and J{\"o}rg Tiedemann. 2017.
\newblock \href {https://www.aclweb.org/anthology/E17-2102} {Continuous
  multilinguality with language vectors}.
\newblock In \emph{Proceedings of the 15th Conference of the {E}uropean Chapter
  of the Association for Computational Linguistics: Volume 2, Short Papers},
  pages 644--649, Valencia, Spain. Association for Computational Linguistics.

\bibitem[{Ponti et~al.(2019{\natexlab{a}})Ponti, O’horan, Berzak, Vuli{\'c},
  Reichart, Poibeau, Shutova, and Korhonen}]{ponti2019modeling}
Edoardo~Maria Ponti, Helen O’horan, Yevgeni Berzak, Ivan Vuli{\'c}, Roi
  Reichart, Thierry Poibeau, Ekaterina Shutova, and Anna Korhonen.
  2019{\natexlab{a}}.
\newblock \href {https://www.mitpressjournals.org/doi/pdf/10.1162/COLI_a_00357}
  {Modeling language variation and universals: A survey on typological
  linguistics for natural language processing}.
\newblock \emph{Computational Linguistics}, 45(3):559--601.

\bibitem[{Ponti et~al.(2018)Ponti, Reichart, Korhonen, and
  Vuli{\'c}}]{ponti-etal-2018-isomorphic}
Edoardo~Maria Ponti, Roi Reichart, Anna Korhonen, and Ivan Vuli{\'c}. 2018.
\newblock \href {https://doi.org/10.18653/v1/P18-1142} {Isomorphic transfer of
  syntactic structures in cross-lingual {NLP}}.
\newblock In \emph{Proceedings of the 56th Annual Meeting of the Association
  for Computational Linguistics (Volume 1: Long Papers)}, pages 1531--1542,
  Melbourne, Australia. Association for Computational Linguistics.

\bibitem[{Ponti et~al.(2019{\natexlab{b}})Ponti, Vuli{\'c}, Cotterell,
  Reichart, and Korhonen}]{ponti-etal-2019-towards}
Edoardo~Maria Ponti, Ivan Vuli{\'c}, Ryan Cotterell, Roi Reichart, and Anna
  Korhonen. 2019{\natexlab{b}}.
\newblock \href {https://doi.org/10.18653/v1/D19-1288} {Towards zero-shot
  language modeling}.
\newblock In \emph{Proceedings of the 2019 Conference on Empirical Methods in
  Natural Language Processing and the 9th International Joint Conference on
  Natural Language Processing (EMNLP-IJCNLP)}, pages 2900--2910, Hong Kong,
  China. Association for Computational Linguistics.

\bibitem[{T{\"a}ckstr{\"o}m et~al.(2013)T{\"a}ckstr{\"o}m, McDonald, and
  Nivre}]{tackstrom-etal-2013-target}
Oscar T{\"a}ckstr{\"o}m, Ryan McDonald, and Joakim Nivre. 2013.
\newblock \href {https://www.aclweb.org/anthology/N13-1126} {Target language
  adaptation of discriminative transfer parsers}.
\newblock In \emph{Proceedings of the 2013 Conference of the North {A}merican
  Chapter of the Association for Computational Linguistics: Human Language
  Technologies}, pages 1061--1071, Atlanta, Georgia. Association for
  Computational Linguistics.

\bibitem[{Tsvetkov et~al.(2016)Tsvetkov, Sitaram, Faruqui, Lample, Littell,
  Mortensen, Black, Levin, and Dyer}]{tsvetkov-etal-2016-polyglot}
Yulia Tsvetkov, Sunayana Sitaram, Manaal Faruqui, Guillaume Lample, Patrick
  Littell, David Mortensen, Alan~W Black, Lori Levin, and Chris Dyer. 2016.
\newblock \href {https://doi.org/10.18653/v1/N16-1161} {Polyglot neural
  language models: A case study in cross-lingual phonetic representation
  learning}.
\newblock In \emph{Proceedings of the 2016 Conference of the North {A}merican
  Chapter of the Association for Computational Linguistics: Human Language
  Technologies}, pages 1357--1366, San Diego, California. Association for
  Computational Linguistics.

\bibitem[{Vastl et~al.(2020)Vastl, Zeman, and Rosa}]{ufal2020sigtyp}
Martin Vastl, Daniel Zeman, and Rudolf Rosa. 2020.
\newblock {Predicting Typological Features in WALS using Language Embeddings
  and Conditional Probabilities: {\'U}FAL Submission to the SIGTYP 2020 Shared
  Task}.
\newblock In \emph{Proceedings of the Second Workshop on Computational Research
  in Linguistic Typology}. Association for Computational Linguistics.

\bibitem[{Velupillai(2012)}]{velupillai2012introduction}
Viveka Velupillai. 2012.
\newblock \emph{An introduction to linguistic typology}.
\newblock John Benjamins Publishing Company Amsterdam, Philadelphia.

\bibitem[{Wada et~al.(2019)Wada, Iwata, and Matsumoto}]{conf/acl/WadaIM19}
Takashi Wada, Tomoharu Iwata, and Yuji Matsumoto. 2019.
\newblock \href {http://dblp.uni-trier.de/db/conf/acl/acl2019-1.html#WadaIM19}
  {{Unsupervised Multilingual Word Embedding with Limited Resources using
  Neural Language Models}}.
\newblock In \emph{ACL (1)}, pages 3113--3124. Association for Computational
  Linguistics.

\bibitem[{Zhang and Barzilay(2015)}]{zhang-barzilay-2015-hierarchical}
Yuan Zhang and Regina Barzilay. 2015.
\newblock \href {https://doi.org/10.18653/v1/D15-1213} {Hierarchical low-rank
  tensors for multilingual transfer parsing}.
\newblock In \emph{Proceedings of the 2015 Conference on Empirical Methods in
  Natural Language Processing}, pages 1857--1867, Lisbon, Portugal. Association
  for Computational Linguistics.

\bibitem[{Zhao et~al.(2020)Zhao, Eger, Bjerva, and
  Augenstein}]{zhao2020inducing}
Wei Zhao, Steffen Eger, Johannes Bjerva, and Isabelle Augenstein. 2020.
\newblock {Inducing Language-Agnostic Multilingual Representations}.
\newblock \emph{arXiv preprint arXiv:2008.09112}.

\end{thebibliography}
\bibliographystyle{acl_natbib}

\end{document}